\documentclass[runningheads]{llncs}

\usepackage[mobile]{eccv}

\usepackage{eccvabbrv}
\usepackage{amsmath}
\usepackage{graphicx}
\usepackage{booktabs}
\usepackage{graphicx}
\usepackage{algorithm}
\usepackage{algorithmic}
\usepackage{epstopdf}

\usepackage[accsupp]{axessibility}  

\usepackage[pagebackref,breaklinks,colorlinks,citecolor=eccvblue]{hyperref}

\usepackage{orcidlink}

\begin{document}

\title{Precise-Physics Driven Text-to-3D Generation} 

\titlerunning{Phy3DGen}

\author{Qingshan Xu\inst{1\star} \and
Jiao Liu\inst{1\star} \and
Melvin Wong\inst{1}\thanks{Equal contribution.} \and \\ 
Caishun Chen\inst{2} \and 
Yew-Soon Ong\inst{1,2}}

\authorrunning{Qingshan~Xu et al.}

\institute{School of Computer Science and Engineering (SCSE), Nanyang Technological University (NTU), Singapore \and
Centre for Frontier AI Research (CFAR), Agency for Science, Technology and Research (A*STAR), Singapore
}

\maketitle

\begin{abstract}
  Text-to-3D generation has shown great promise in generating novel 3D content based on given text prompts. However, existing generative methods mostly focus on geometric or visual plausibility while ignoring precise physics perception for the generated 3D shapes. This greatly hinders the practicality of generated 3D shapes in real-world applications. In this work, we propose Phy3DGen, a precise-physics-driven text-to-3D generation method. By analyzing the solid mechanics of generated 3D shapes, we reveal that the 3D shapes generated by existing text-to-3D generation methods are impractical for real-world applications as the generated 3D shapes do not conform to the laws of physics. To this end, we leverage 3D diffusion models to provide 3D shape priors and design a data-driven differentiable physics layer to optimize 3D shape priors with solid mechanics. This allows us to optimize geometry efficiently and learn precise physics information about 3D shapes at the same time. Experimental results demonstrate that our method can consider both geometric plausibility and precise physics perception, further bridging 3D virtual modeling and precise physical worlds.  
\end{abstract}

\section{Introduction}
\label{sec:intro}
Text-to-3D generation can synthesize novel 3D content based on text prompts. It is critical for a variety of applications, including personalized customization, film-making,  robotics simulation, gaming, and so on. Recently, text-to-3D generation~\cite{poole2022dreamfusion,lin2023magic3d,tang2023dreamgaussian} have achieved promising results with the development of generative models (\eg, generative adversarial networks~\cite{goodfellow2020generative} and diffusion models~\cite{ho2020denoising,rombach2022high}) and 3D representations (\eg, DeepSDF~\cite{park2019deepsdf} and neural radiance fields~\cite{mildenhall2021nerf}). Existing text-to-3D generation methods mostly focus on the geometric or visual realism of generated 3D shapes. However, since these methods do not incorporate precise physics information, most 3D shapes generated are ill-defined and impractical for real-world applications.

\begin{figure}
    \centering
    \includegraphics[width=\linewidth]{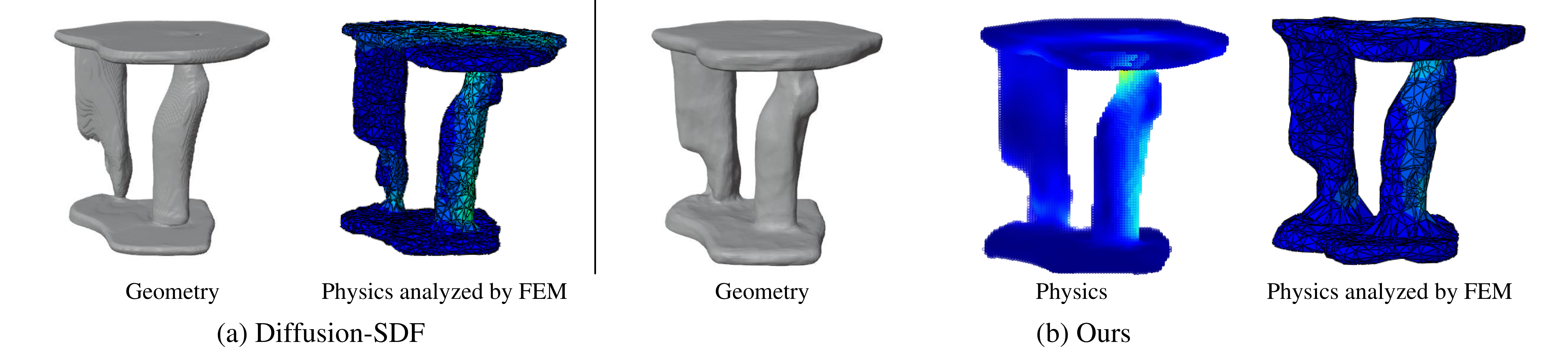}
    \caption{\textbf{Motivation.} By applying distributed force on the object top, we analyze the solid mechanics of 3D shapes generated by text-to-3D generation methods by FEM. For physics, the lighter the color values, the higher the stress levels. We observe that the generated 3D shape by Diffusion-SDF~\cite{li2023diffusion} will experience notably higher stress in some regions, demonstrating that the generated geometry is fragile. In contrast, the stress distribution of our generated 3D shape is more uniform as the generated geometry conforms to the physical laws. Moreover, our learned physics information is comparable to the physics obtained by FEM.}
    \label{fig:teaser}
\end{figure}

Although existing text-to-3D generation methods can learn diverse and plausibly reasonable shape priors from large amounts of data based on generative models, these 3D shapes generated do not conform to the laws of physics as there is no guarantee that these methods can learn precise physical information from these data. {In this work, we treat the object described by the 3D geometry as a solid and incorporate basic solid mechanics properties into our analysis. Specifically, we conform the 3D geometry to linear elasticity~\cite{bertram2015solid} behavior.} {In \cref{fig:teaser}, we analyze the solid mechanics of a 3D shape generated by Diffusion-SDF using the conventional Finite Element Method (FEM)~\cite{zienkiewicz2000finite}. By applying distributed force on the tabletop, we can observe that stress (quantified by the von-Mises stress) spreads across the entire geometry, with certain regions experiencing notably higher stress levels than others. Typically, these concentrated high-stress regions are potential areas of failure due to external forces. Therefore, minimizing the maximum stress and uniformly distributing the stress throughout the geometry can reduce susceptibility to failure. This is highly valuable in engineering design. This paper aims to achieve geometry generation and optimization, considering not only computer vision but also physics. This approach ensures that the obtained geometry not only satisfies visual preferences but also meets engineering requirements to a certain extent.}

However, it is challenging to incorporate precise physics information into text-to-3D generation methods. On the one hand, in order to optimize geometry with physics, a differentiable physics solver is expected to compute the solid mechanics for the intermediate 3D shapes on the fly. However, conventional FEM solvers, as offline simulators, are discrete and, more importantly, non-differentiable. Therefore, it is hard to apply FEM solvers to efficiently optimize 3D shapes directly. On the other hand, recent Physics Informed Neural Networks (PINN) provide a new direction to solve Partial Differential Equations (PDE) describing the laws of physics. However, they are still under-explored in computational solid mechanics and their precision is poor for complex geometries. This further prevents precise physics perception for generative 3D methods.        

To address the aforementioned challenges, we propose a precise physics-driven text-to-3D generation method to inject the physical laws into the generative 3D modeling. Specifically, our method contains two stages. In the first stage, we initialize a 3D shape based on 3D diffusion models and convert it into implicit geometry representations, Signed Distance Function (SDF)~\cite{park2019deepsdf}. This allows our approach to be flexibly combined with existing generative methods. In the second stage, we design a data-driven differentiable physics layer to enable geometry and physics optimization at the same time. Our differentiable physics layer is parameterized by neural networks and used to learn the solid mechanics of generated 3D shapes on demand when optimizing geometries. To guarantee the precision of the feedback physics information, we leverage the FEM results computed for the initial generated 3D shape to initialize our differentiable physics layer. We introduce a relaxed geometry loss and a series of physics losses during training to guide the optimization. Moreover, we design an alternating training strategy to learn geometry and physics information collaboratively. In this way, our method can simultaneously optimize geometry and learn the physics of generated 3D shapes, considering both geometric plausibility and precise physics perception.

Overall, our contributions are summarized as follows:
\begin{itemize}
    \item We propose a precise-physics-driven text-to-3D generation method, called Phy3DGen, to incorporate the physical laws into generated 3D shapes. This bridges 3D virtual modeling and precise physical worlds.
    \item We design a data-driven differentiable physical layer to learn the physics information during training effectively. Moreover, an alternating training strategy is introduced to stabilize the learning of physics. Therefore, our method can collaboratively optimize geometry and physics flexibly and reliably. 
    \item Experiments demonstrate that our method can improve geometric or visual preferences without requiring extra training data and manual interaction and endow generated 3D shapes with precise physical perception capabilities.
\end{itemize}

\section{Related Work}

\subsection{Text-to-3D Generation} Recently, text-to-3D generation has achieved significant success. On the one hand, some generation methods train their models on text-3D pairs and can directly generate 3D shapes from texts without requiring any optimization at inference~\cite{sanghi2022clip,mittal2022autosdf,li2023diffusion,nichol2022point,jun2023shap}. CLIP-forge~\cite{sanghi2022clip} uses renderings of shapes with pre-trained image-text joint embedding to learn shapes conditioned on texts. AutoSDF~\cite{mittal2022autosdf} explores an auto-regressive prior for 3D shape generation based on a discretized SDF autoencoder. Recent works employ diffusion models to learn a probabilistic mapping from the text to 3D shapes, such as Diffusion-SDF~\cite{li2023diffusion}, Point-E~\cite{nichol2022point} and Shape-E~\cite{jun2023shap}. On the other hand, some methods lift 2D diffusion models for text-to-3D generation~\cite{poole2022dreamfusion,lin2023magic3d,tang2023dreamgaussian,wang2024prolificdreamer}. These methods, such as DreamFusion~\cite{poole2022dreamfusion} and DreamGaussian~\cite{tang2023dreamgaussian}, leverage pre-trained text-to-image diffusion models as the guidance to optimize 3D representations, including neural radiance fields (NeRF)~\cite{mildenhall2021nerf,barron2021mip,muller2022instant,fu2022geo} and Gaussian Splatting~\cite{tang2023dreamgaussian}. Therefore, these methods do not require any 3D data. However, whether these generation methods train their models on 3D data or rely on text-to-image priors trained on 2D data, there is no guarantee that they can perceive precise physics information from large amounts of data. This work shows that our method can learn precise physics properties for the generated 3D shapes during training. 

\subsection{Physics-aware 3D Generation} There exist several physics-aware 3D generation methods~\cite{mezghanni2021physically,mezghanni2022physical,liu2023few,wang2019physics}. These methods utilize either offline simulations or online simulations to ensure the physical validity of generated shapes. \cite{mezghanni2021physically} pre-trains a surrogate network using offline simulations to predict stability to inform the generative learning. \cite{mezghanni2022physical,liu2023few} design online simulation layers to provide physical guidance for the generative learning. All these methods do not require to learn physics feedback during shape learning. Different from them, our method collaboratively learns geometry and physics during training. 

\subsection{Physics-Informed Machine Learning}
{
The simulation and modeling of physical systems have long played a crucial role in various scientific and engineering domains \cite{heermann1990computer}. Traditionally, the FEM has been instrumental in these tasks. However, recent advancements in deep neural networks have prompted researchers to explore novel applications in these domains \cite{raissi2018deep}. Among these methods, PINNs have garnered significant attention \cite{raissi2019physics}. PINNs stand out due to their ability to integrate governing equations of mechanics, often expressed through PDEs, into the learning objectives of neural networks. This integration ensures the network outputs align with the fundamental governing physics. Consequently, there has been a proliferation of studies on PINNs in recent years \cite{deng2023physical,chiu2022can,mao2020physics,wang2023expert}. While several methods have been proposed, the majority of research on PINNs primarily focuses on simulating physical systems. There is a limited number of discussions on optimization based on PINNs, such as physics-informed inverse design~\cite{lu2021physics}, PDE constrained optimization~\cite{hao2022bi}, and PINN-assisted topology optimization~\cite{jeong2023complete,yin2023dynamically}. However, most of these works concentrate on optimizing simple functions or 2D geometries. None of them address the optimization of 3D geometries.
}

\section{Method}   

\begin{figure}[t]
    \centering
    \includegraphics[width=\linewidth]{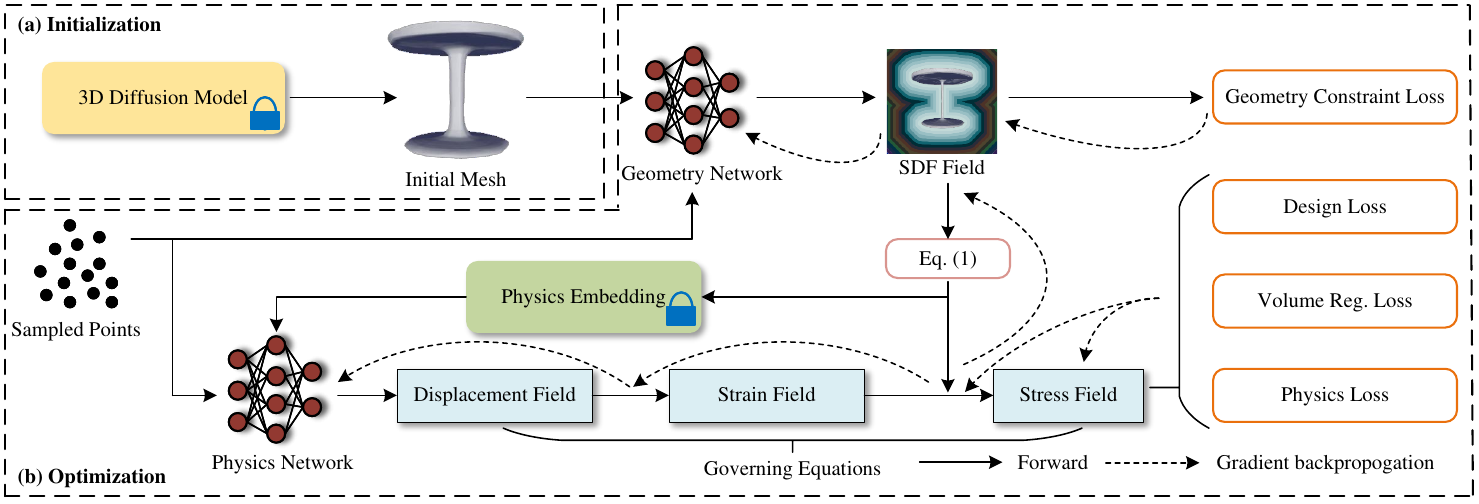}
    \caption{\textbf{Overall framework of our method.} Our method consists of two stages, initialization and optimization. We employ a 3D diffusion model to generate an initial mesh in the initialization stage. Then, the mesh is used to initialize the geometry network. Physics embedding is first computed to initialize the physics network in the optimization stage. Then, we sample 3D points to compute their geometry (SDF) and physics (displacement, strain, stress) properties. In this way, a total loss composed of geometry constraint loss, design loss, volume regularization loss, and physics loss is calculated to optimize the geometry and physics networks simultaneously.}
    \label{fig:framework}
\end{figure}

Given a text describing our target object, we aim to generate a 3D shape that simultaneously satisfies geometric plausibility and precise physics perception. For this purpose, we propose a precise-physics-driven text-to-3D generation method. This section first gives an overview of our whole framework in \cref{sec:overview}. Then, we describe the geometry representations and initialization in \cref{sec:geometry}. We elaborate on our physics representations and governing equations used to model the physical laws in \cref{sec:physics}. In \cref{sec:co-training}, we detail our {precise-physics embedding and training process}.

\subsection{Overview}\label{sec:overview}

 \cref{fig:framework} depicts the overall framework of our method. It is a two-stage pipeline, including initialization with diffusion-based methods and optimization with our designed differentiable physics layer. In the initialization stage, we generate an initial 3D shape by leveraging diffusion-based text-to-3D generation methods and convert it into a neural implicit representation, i.e., SDF, which is parameterized by neural networks. In the optimization stage, we design a data-driven differentiable physics layer to represent physics by neural networks implicitly. Then, the implicit physics representations are initialized by the FEM results computed for the initial 3D shape and our designed physics losses. On this basis, we further introduce relaxed geometry losses and an alternating training strategy to optimize the 3D geometry and its corresponding physics information collaboratively.      

\subsection{Geometry Representations and Initialization}\label{sec:geometry}

\noindent\textbf{Geometry Representations.} Our method consists of two key representations, geometry and physics. In order to allow our geometry to support physics perception, the geometry representation should have the following properties: 
\begin{itemize}
    \item \emph{Differentiability of geometry}: The geometry representation should be differentiable to allow for end-to-end back-propagation during training.
    \item \emph{Scalability of geometry}: The geometry representation should be scalable to allow for fine-grained sampling in the entire 3D space.
    \item \emph{Discriminability of interior and exterior}: For physics, the solid mechanics will work on geometry surfaces and interior. Therefore, the geometry representation should make it easy to distinguish between interior and exterior. 
\end{itemize}

To this end, we employ an implicit SDF~\cite{park2019deepsdf} encoded by neural networks to represent our geometry. It is naturally differentiable and can be evaluated at different resolutions. For any 3D point $\textbf{x}$, its SDF value $\hat{f}_S(\textbf{x})$ represents the signed distance to the geometry surfaces. That is, if the output SDF is a negative value, the corresponding point will be outside the object; otherwise, the corresponding point will be inside. Therefore, it is easy for point $\textbf{x}$ to distinguish between interior and exterior with its SDF sign. 

{As physics calculations are contingent upon geometry, it's imperative to find a representation that the physics network can easily understand. Inspired by topology optimization \cite{bendsoe2013topology}, we opt to convert the SDF to the density field function.} This is achieved as follows:
\begin{equation}
    \hat{\rho}(\textbf{x}) = \text{Sigmoid}(\frac{\hat{f}_{S}(\textbf{x})}{\tau}),
\end{equation}
where $\tau$ is the temperature hyper-parameter. Different from the SDF $\hat{f}_{S}(\textbf{x})$, we follow the same assumption with ``Solid Isotropic Material with Penalisation''~\cite{bendsoe1989optimal,bendsoe1999material}, i.e., the density field function $\hat{\rho}(\textbf{x})$ will directly influence the physical properties of the material in the entire 3D space, such as Young's module and Possion's ratio~\cite{zienkiewicz2000finite}. This definition allows incorporating the geometry information into the physics layer.

\noindent\textbf{Geometry Initialization.} To initialize our geometry representations, we use a 3D diffusion models to generate an initial 3D shape $\mathcal{M}$ based on the text prompts $\mathcal{T}$. In general, 3D diffusion models will output meshes to represent 3D shapes because mesh representations provide a clear topological structure, making them convenient for operations and analysis. To allow our method for the subsequent optimization with SDF, we convert the mesh $\mathcal{M}$ to an initial SDF $f_{init}$ by training a geometry network $G_\theta$ parameterized by $\theta$. This geometry network transfers the shape prior information from 3D diffusion models, thus we can focus on optimizing this geometry network in our subsequent stage. 

\subsection{Physics Representations and Governing Equations}\label{sec:physics} 

\noindent\textbf{Physics Quantities.}  {To describe the solid mechanics of geometry, we consider three fundamental physical quantities within the context of physics~\cite{bertram2015solid}:
\begin{itemize}
\item \textit{Displacement}: Representing the shift of any point within a geometric structure, displacement is characterized by a 3D vector $\textbf{u}(\textbf{x}) = (u_1(\textbf{x}), u_2(\textbf{x}), u_3(\textbf{x}))$.
\item \textit{Strain}: Strain serves as a measure of deformation, depicting the relative displacement between points within a geometric body. Typically, strain at a point is denoted by a second-order tensor $\boldsymbol{\epsilon}(\textbf{x}) = \{ (\epsilon_{ij}(\textbf{x}))| i,j = 1,2,3 \}$.
\item \textit{Stress}: Stress encapsulates the internal forces exerted by neighboring points within a continuous geometric entity. The Cauchy stress tensor, often employed to describe stress, is denoted as $\boldsymbol{\sigma}(\textbf{x}) = \{ (\sigma_{ij}(\textbf{x}))| i,j = 1,2,3 \}$.
\end{itemize}
}

{It is essential to recognize that the applicability of physics is often contingent upon specific external environmental circumstances, commonly described by \textit{boundary conditions}. In this study, the boundary conditions for a particular geometry are established based on typical usage scenarios of the object represented by said geometry. For instance, in the case of a table, a force may be applied to one side of the object in accordance with practical requirements. }

\noindent\textbf{Governing Equations.} {Governing equations serve as a foundational set of equations, often comprising ordinary or partial differential equations, to articulate the behavior of a physical system. As mentioned in \cref{sec:intro}, in this work, our focus lies primarily on the classical linear elastic system, which aptly characterizes elastic bodies under conditions of minimal deformation. The following equations delineate this system:
\begin{equation}\label{eq:gover_eqn1}
\boldsymbol{\nabla} \cdot \boldsymbol{\sigma} + \textbf{F} = \textbf{0},
\end{equation}

\begin{equation}\label{eq:gover_eqn2}
\boldsymbol{\epsilon} = \frac{1}{2} [ \boldsymbol{\nabla} \textbf{u} + (\boldsymbol{\nabla} \textbf{u})^{\intercal} ],
\end{equation}

\begin{equation}\label{eq:gover_eqn3}
\boldsymbol{\sigma} = \textbf{C} : \boldsymbol{\epsilon}
\end{equation}
Here, $\textbf{F}$ denotes body force per unit mass, ``$:$'' is the double dot product, and $\textbf{C}$ is the fourth-order stiffness tensor. Note that, in this paper, we assume that a shape is made by the isotropic material. At this time, all elements in $\textbf{C}$ are decided by only two parameters, Young's module and Poisson's ratio. Typically, these two parameters are determined by the material properties of the geometric body under consideration. The details can be found in~\cite{zienkiewicz2000finite}. For the sake of simplicity and generality within this study, we set these parameters as 1 and 0.3, respectively. 
The ensuing boundary conditions supplement these governing equations:

\begin{equation}\label{eq:gover_eqn4}
\begin{aligned}
& \textbf{u} = \bar{\textbf{u}}, \ \ \ \textbf{x} \in \Gamma_u, \\
& \boldsymbol{\sigma} \cdot \textbf{n} = \bar{\textbf{F}}, \ \ \ \textbf{x} \in \Gamma_f. \\
\end{aligned}
\end{equation}
Here, $\bar{\textbf{u}}$ and $\bar{\textbf{F}}$ represent the already known displacement and force, at the corresponding boundaries $\Gamma_u$ and $\Gamma_t$, while $\textbf{n}$ denotes the unit outward normal vector on the relevant boundaries $\Gamma_f$.}

\noindent\textbf{Physics Representations based on Neural Networks.}
{
Similar to PINNs, in our physics layer, we employ a neural network $U_{\phi}$ parameterized by $\phi$ to predict the displacement $\hat{\textbf{u}}(\textbf{x})$ at any point $\textbf{x}$ within the geometry. Subsequently, utilizing automatic differentiation based on \cref{eq:gover_eqn3}, we obtain the predicted strain tensor $\hat{\boldsymbol{\epsilon}}(\textbf{x})$. According to  \cref{eq:gover_eqn2}, we compute the predicted Cauchy stress tensor $\hat{\boldsymbol{\sigma}}(\textbf{x})$.
}

\subsection{Precise-Physics Embedding and Training}\label{sec:co-training}

\noindent\textbf{Precise-Physics Embedding.}
{To enhance the physics guidance on geometry optimization, a high-precision simulator capable of providing accurate physical insights is essential. FEM stands out as a classical and reliable simulator renowned for its precision in capturing physical phenomena, especially in simple systems like linear elastic setups. However, FEM-based simulators often lack differentiability, which poses a challenge when integrating them into online geometry optimization processes. Physics-informed machine learning has recently emerged as a promising alternative to physical simulations. Their differentiable nature, stemming from neural network architecture, facilitates seamless integration into online geometric optimization workflows. Nonetheless, physics-informed machine learning is still evolving, and current implementations encounter difficulties in delivering highly precise simulations for complex 3D geometries. To this end, we introduce a technique called precise-physics embedding.}

{The underlying concept of precise-physics embedding is: leveraging the differentiability inherent in neural networks while enhancing the precision of physics predictions. This is achieved by integrating FEM data into the training process of our physics layer. By combining the differentiability of neural works with the high precision provided by FEM data, we can effectively drive geometric optimization using differentiable properties while ensuring accurate physics guidance throughout the optimization process.}

\noindent\textbf{Precise-Physics Driven Pretraining.} 
{During the pretraining process, we keep the initial geometry unchanged and focus mainly on obtaining its precise physical information by utilizing a neural network. The whole process of the precise-physics driven pretraining contains the following steps:
\begin{itemize}
    \item \textit{Data Generation}: We assume that a regular geometric space $\Omega^{+}$ (i.e., a cube) can adequately encompass the original geometry $\Omega$. Consequently, we sample a set of points $\mathcal{D}_{\Omega^{+}} = \{ (\textbf{x}^{(l)}) \}_{l=1}^{N_{\Omega^{+}}}$ within $\Omega^{+}$ to evaluate whether the predictions provided by the physics layer satisfies the governing equations. Additionally, we require two other sets of sample points, denoted as $\mathcal{D}_{\Gamma_{u}} = \{ (\textbf{x}_{\Gamma_{u}}^{(l)}) \}_{l=1}^{N_{\Gamma_{u}}}$ and $\mathcal{D}_{\Gamma^{+}_{f}} = \{ (\textbf{x}_{\Gamma_{f}}^{(l)}) \}_{l=1}^{N_{\Gamma^{+}_{f}}}$. These sets contain sample points on the traction and traction-free boundaries of $\Omega^{+}$, respectively. Furthermore, we simulate the physics for the initial geometry using FEM, resulting in a precise-physics dataset $\mathcal{D}_{fem} = \{ \textbf{x}_{fem},\textbf{u}_{fem},\boldsymbol{\sigma}_{fem} \} = \{ (\textbf{x}_{fem}^{(l)},\textbf{u}_{fem}^{(l)},\boldsymbol{\sigma}_{fem}^{(l)}) \}_{l=1}^{N_{fem}}$. This dataset serves to enhance the pretraining of the physics layer and provide better guidance for the optimization process.

    \item \textit{Geometry Incorporation}: We integrate geometry information into the physics layer inspired by topology optimization techniques~\cite{yin2023dynamically}. Specifically, it is achieved as follows:
    \begin{equation}\label{eq:pred_sigma}
        \begin{aligned}
            & \boldsymbol{\hat{\sigma}} = \hat{\rho} \textbf{C} : \boldsymbol{\epsilon}, 
        \end{aligned}
    \end{equation}
    and the equilibrium equation \cref{eq:gover_eqn1} is modified as:
    \begin{equation}\label{eq:modegover_eqn1}
        \begin{aligned}
            \boldsymbol{\nabla} \cdot \hat{\boldsymbol{\sigma}} + \hat{\rho} \textbf{F} = \textbf{0}.
        \end{aligned}
    \end{equation}
    Thus, the predicted Cauchy stress tensor $\hat{\boldsymbol{\sigma}}$ can also be provided by combining the predictions from both the geometry network and the physics layer.

    \item \textit{Pretraining}: To ensure that the physics layer provides highly accurate predictions for the physics, it must satisfy both the governing equations and the simulation results obtained from FEM. Therefore, the loss function for the pretraining process is defined as follows:
    \begin{equation}\label{eq:depinn_loss}
        \begin{aligned}
            L_{pl} = w_{pde} L_{pde} + w_{bc} L_{bc^{+}} + w_{fem} L_{fem},
        \end{aligned}
    \end{equation}
    \begin{equation}\label{eq:depde_loss}
        \begin{aligned}
            L_{pde} = || \boldsymbol{\nabla} \cdot \hat{\boldsymbol{\sigma}} + \hat{\rho} \textbf{F} ||_{\Omega^{+}}^{2},
        \end{aligned}
    \end{equation}
    \begin{equation}\label{eq:debc_loss}
        \begin{aligned}
            L_{bc} = || \hat{\textbf{u}} - \bar{\textbf{u}} ||_{\Gamma_u}^{2} +  || \hat{\boldsymbol{\sigma}} \cdot \textbf{n} - \bar{\textbf{F}} ||_{\Gamma^{+}_f}^{2},
        \end{aligned}
    \end{equation}

     \begin{equation}\label{eq:defem_loss}
        \begin{aligned}
            L_{fem} = || \hat{\textbf{u}}_{fem} - {\textbf{u}}_{fem} ||^{2} +  || \hat{\boldsymbol{\sigma}}_{fem} \cdot \textbf{n} - \boldsymbol{\sigma}_{fem} ||^{2},
        \end{aligned}
    \end{equation}
    where $\Gamma_{f}^{+}$ contains all of the sample points on the traction force boundaries of $\Omega$ and the traction-free boundaries of $\Omega^{+}$, and $\hat{\textbf{u}}_{fem}$ and $\hat{\boldsymbol{\sigma}}_{fem}$ are predictions provided by the physics layer for the sample points in $\mathcal{D}_{fem}$. In terms of the above loss functions, $L_{pde}$ is estimated based on $\mathcal{D}_{\Omega^{+}}$, $L_{bc}$ is estimated based on $\mathcal{D}_{\Gamma_{u}}$ and $\mathcal{D}_{\Gamma^{+}_{f}}$, and $L_{fem}$ is calculated based on $\mathcal{D}_{fem}$. These loss functions aim to make the predictions of the physics layer satisfy the PDE, the boundary conditions, and the precise physics data, respectively.
\end{itemize}
}
\noindent\textbf{Precise-Physics Driven Co-training.}
{ 
We introduce a co-training framework for iterative geometry refinement to optimize the geometry while incorporating physical information. During this process, we aim to enhance the adherence of the geometry to objective physical laws while avoiding drastic changes that deviate from the intended concept and maintaining aesthetic qualities. Therefore, we introduce the following loss functions in our co-training process:
}
{
\begin{itemize}
    \item {\textit{Design Loss}: Unlike conventional physics-informed machine learning, our objective extends beyond accurately computing the physics fields; we aim to achieve optimal geometry. Hence, we need to devise an indicator to assess the quality of a given geometry. The indicator we consider is based on achieving uniform stress distribution throughout the geometry, thereby minimizing stress concentration. Accordingly, the design loss is defined as follows:
        \begin{equation}\label{design_loss}
            \begin{aligned}
                L_{design} = \max_{\textbf{x} \in \Omega^{+}} \{\hat{\rho} \hat{\sigma}_{vm}\}
                - \frac{\int_{\textbf{x} \in \Omega^{+}} \hat{\rho} \hat{\sigma}_{vm} \text{d}\textbf{x}}{\int_{\textbf{x} \in \Omega^{+}} \hat{\rho} \text{d}\textbf{x}} ,
            \end{aligned}
        \end{equation}
    where $\hat{\sigma}_{vm}$ is the approximated von-Mises stress, which can be calculated as
        \begin{equation}\label{vm_stress}
            \begin{aligned}
                \hat{\sigma}_{vm} = \sqrt{\frac{3}{2} \left( \hat{\boldsymbol{\sigma}} - \frac{\text{tr}(\hat{\boldsymbol{\sigma}})}{3} \textbf{I} \right) : \left( \hat{\boldsymbol{\sigma}} - \frac{\text{tr}(\hat{\boldsymbol{\sigma}})}{3} \textbf{I} \right)}.
            \end{aligned}
        \end{equation}}
    In \cref{vm_stress}, $\textbf{I}$ is an identity matrix, and \cref{design_loss} can be estimated based on $\mathcal{D}_{\Omega^{+}}$.

    \begin{algorithm}[t]
    	\caption{Precise-Physics Driven Co-training}\label{Alg:Cotraining}
        \renewcommand{\algorithmicrequire}{\textbf{Input:}} 
    	\renewcommand{\algorithmicensure}{\textbf{Output:}}
    	\begin{algorithmic}[1]
            \REQUIRE $\mathcal{D}_{\Omega^{+}}$, $\mathcal{D}_{\Gamma_{u}}$, $\mathcal{D}_{\Gamma^{+}_{f}}$, $\mathcal{D}_{fem}$, $\mathcal{D}_{gc}$, $t$, and $epoch_{max}$;
    		\ENSURE $G_{\theta}$, $U_{\phi}$;
    	    \STATE $epoch = 0$;
            \FOR {$epoch < epoch_{max}$}
            \IF{$epoch \% t == 0$}
            \STATE Freeze the parameters of $U_{\phi}$;
            \STATE Calculate $L_{cotrain}$ based on $\mathcal{D}_{\Omega^{+}}$, $\mathcal{D}_{\Gamma_{u}}$, $\mathcal{D}_{\Gamma^{+}_{f}}$, $\mathcal{D}_{fem}$, and $\mathcal{D}_{gc}$;
            \STATE $\theta \leftarrow \theta - \alpha \cdot \nabla_{\theta} L_{cotrain}$;
            \ELSE
            \STATE Freeze the parameters of $G_{\theta}$;
            \STATE Calculate $L_{pl}$ by utilizing $\mathcal{D}_{\Omega^{+}}$, $\mathcal{D}_{\Gamma_{u}}$, $\mathcal{D}_{\Gamma^{+}_{f}}$, and $\mathcal{D}_{fem}$;
            \STATE $\phi \leftarrow \phi - \beta \cdot \nabla_{\phi} L_{pl}$;
            \ENDIF
            \ENDFOR
            \STATE Output the SDF function.
    	\end{algorithmic}
    \end{algorithm}
    
    \item \textit{Geometry Constraint Loss}: The geometry constraint loss is designed to avoid excessive deformation of geometry. This is achieved by utilizing a dataset $\mathcal{D}_{gc} = \{(\textbf{x}^{(i)}_{gc},\rho_{gc}^{(i)})\}_{i=1}^{N_{gc}}$ sampled outside the envelope of the geometry. We assume that, for these points, the corresponding density values are kept unchanged. Thus, the geometry constraint loss can be calculated as:
    \begin{equation}
        \begin{aligned}
            L_{gc} = \frac{1}{N_{gc}} \sum_{i=1}^{N_{gc}} \left(\hat{\rho} (\textbf{x}_{gc}^{(i)})  - \rho_{gc}^{(i)} \right)^2.
        \end{aligned}
    \end{equation}

    \item \textit{Volume Regularization Loss}: Similar to the geometry constraint loss, the volume regularization loss is to constrain the volume regularization of the optimized object to the target value. Its meaning is to ensure that the overall mass of the geometry is equal to or does not exceed a preset value. In this document, the volume regularization loss is defined as follows:
        \begin{equation}
            \begin{aligned}
                L_{vr} =\max \{ \int_{\textbf{x} \in \Omega^{+}} \hat{\rho} \text{d}\textbf{x} - M_{v}, 0 \}.
            \end{aligned}
        \end{equation}
    where $M_{v}$ is the maximum total volume. This can be estimated based on $\mathcal{D}_{\Omega^{+}}$.
    
    \item \textit{Eikonal Loss}: To ensure the smoothness of the geometric surface, we also introduced the eikonal regularization, which can be described as follows:
    \begin{equation}
        \begin{aligned}
            L_{eikonal} = (|| \nabla \hat{f}_{S}|| - 1)^2.
        \end{aligned}
    \end{equation}
\end{itemize}

In addition to the aforementioned loss functions, $L_{pde}$, $L_{bc}$, and $L_{fem}$ are also incorporated into the co-training process. It's important to note that the dataset provided by FEM corresponds to the initial geometry. The associated physics values will also change as the geometry evolves during the co-training process. However, we assume that the changes in geometry are not substantial (we also employ $L_{gc}$ and $L_{vr}$ to achieve this), and the FEM data for the initial geometry can still offer guidance for optimization. Therefore, $L_{fem}$ remains employed throughout the co-training process.

The pseudo-code of the co-training process is shown in \textbf{Algorithm 1}. In this process, we alternately optimize two integrated loss functions $L_{pl}$ and $L_{cotrain}$. Specifically, $L_{pl}$ is the same as \cref{eq:depinn_loss}, and $L_{cotrain}$ is calculated as follows:
    \begin{equation}\label{eq:cotrain}
        \begin{aligned}
            L_{cotrain} = w_{design} L_{design} + w_{gc} L_{gc} + w_{vr} L_{vr} + w_{eikonal} L_{eikonal},
        \end{aligned}
    \end{equation}
where $w_{design}$, $w_{gc}$, $w_{vr}$, and $w_{eikonal}$ are the weights for different loss functions. Note that our focus lies on the convergence of $L_{pl}$ to ensure that the PINN can provide sufficiently accurate physical predictions, thereby ensuring that the physics can offer correct guidance for geometry optimization. Consequently, for every $t$ epoch, we optimize $L_{cotrain}$ for one step. Further details are outlined in \textbf{Algorithm 1}, and $\alpha$ and $\beta$ are the learning rates in the optimization process. 
}

\section{Experiments}

\subsection{Implementation Details}

Our method is implemented by PyTorch~\cite{paszke2019pytorch}. The 3D diffusion model used in our method to obtain the initial 3D mesh is Diffusion-SDF~\cite{li2023diffusion}. This model mainly generates two object types: chair and table. In addition, we will also validate our method using another 3D diffusion model, Shap-E~\cite{jun2023shap}.

Our geometry network $G_\theta$ is modeled by an 8-layer multi-layer perceptron (MLP) with 256 hidden units. It is initialized by the geometric
initialization~\cite{atzmon2020sal}. To transfer the shape priors from 3D diffusion models to the geometry network, we follow DeepSDF~\cite{park2019deepsdf} to train our geometry network. The geometry network is trained for 50 epochs, with each epoch comprising 200 training steps. A batch size of 16,384 SDF samples is employed, and the learning rate is set to 5e-4.

Our physics network $U_\phi$ is modeled by a 6-layer MLP with 125 hidden units. To train the physics networks efficiently and effectively, a dense-sparse strategy is employed to partition computation regions into ``dense'' regions where physics precision is crucial, and ``sparse'' regions where physics precision can be relaxed (\ie regions distant from the mesh). The dense-sparse ratio is set at 0.3, with the dense regions being defined as regions where $f_{init}(\textbf{x})$ {\textless} 0.10. The physics network is trained for 10,000 epochs with a learning rate 5e-3.

\begin{figure}[t]
    \centering
    \includegraphics[width=\linewidth]{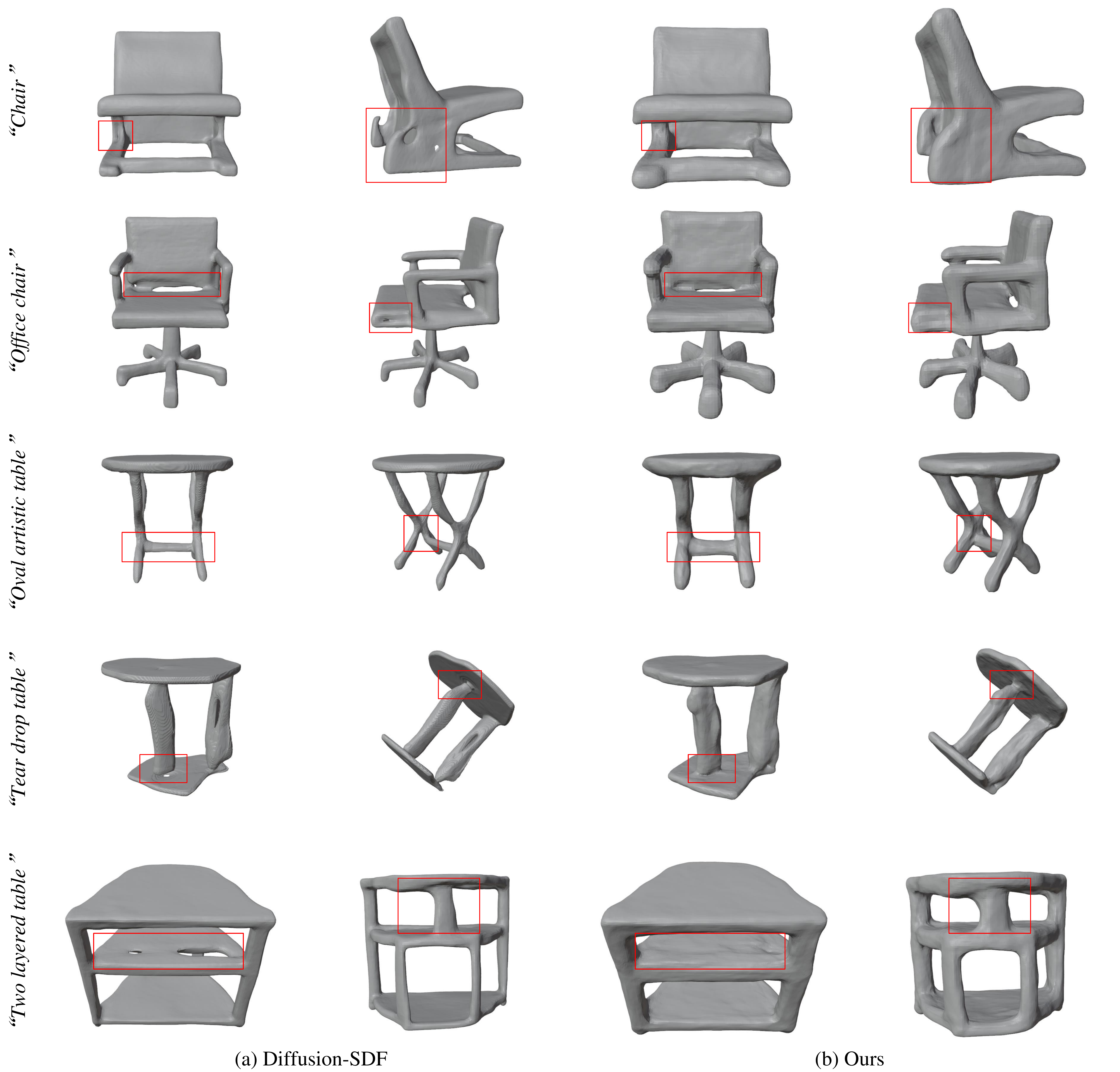}
    \caption{\textbf{Qualitative geometry comparison.}}
    \label{fig:geo_comp}
\end{figure}

The co-training process spans 500 epochs, with a learning rate of 1e-3 set for the physics network and 1e-6 for the geometry network. For the \cref{eq:depinn_loss}, $w_{pde}$, $w_{bc}$, and $w_{fem}$ are set to 1, 50, and 1, respectively. For the \cref{eq:cotrain}, $w_{design}$, $w_{gc}$, $w_{vr}$ and $w_{eikonal}$ are set to 25, 100,000, 10,000 and 10, respectively. The $t$ in \cref{Alg:Cotraining} is set to 10. In terms of the FEM data, we provide the displacement and stress values for the tetrahedral with 7000 nodes. All these data are obtained using Abaqus~\cite{barbero2023finite}, version 2016. 

\subsection{Geometry Comparisons}

We show the geometry comparison results with Diffusion-SDF in \cref{fig:geo_comp}. By giving a text description, Diffusion-SDF and our method will generate the corresponding object. We observe that both Diffusion-SDF and our method can generate visually plausible 3D shapes. Moreover, our method can generate more complete and sturdy objects. For example, when generating a ``two layered table'', Diffusion-SDF generates a table with some holes. By incorporating precise physics into our method, our method needs to consider a more uniform stress distribution. Therefore, our method can generate a more complete 3D shape. When generating an object containing some complex connections, Diffusion-SDF produces fragile connections, like the connections in the red boxes of the ``tear drop table''. Moreover, sometimes, Diffusion-SDF is unable to place some connections to correct positions, such as the ``oval artistic table''. Differently, thanks to our precise physics perception, our method can generate sturdy connections in the correct positions. In addition, the physics information helps our method generate more symmetric 3D shapes, like the ``chair''. Therefore, by incorporating precise physics into 3D generative methods, our method can generate higher-quality geometries, considering both visual realism and practical needs.

\subsection{Analysis of Physics Conformity}
\begin{table}[b]
\centering
\caption{The maximal von-Mises stress in the shapes generated by the Diffusion-SDF and our method (lower is better). All of the results are obtained by FEM.}\label{table:fem}
\begin{tabular}{l|c|c}
\hline
                    & Diffusion-SDF~\cite{li2023diffusion} & Ours    \\
\hline
Chair               & 15.6467       & \textbf{5.8478}  \\
Office Chair        & 40.5651       & \textbf{27.7762} \\
Oval Artistic Table & 31.0362       & \textbf{9.7501}  \\
Round Table         & 6.7947        & \textbf{2.1048}  \\
Tear Drop Table     & 16.7696       & \textbf{8.0138}  \\
Two Layer Table     & 10.8793       & \textbf{2.7844}  \\
\hline
\end{tabular}
\end{table}

{
In this subsection, we comprehensively evaluate the generated shapes from a physics perspective. To achieve this, we perform FEM simulations on shapes generated by both the Diffusion-SDF and our proposed approach. In these simulations, we impose uniformly distributed pressure loading on the upper surface of each shape, mimicking real-world scenarios such as chairs or tables under typical usage conditions. Consequently, we obtain a stress field for each shape, from which we calculate the maximal von-Mises stress and record the results in Table~\ref{table:fem}. Analysis of the data in Table~\ref{table:fem} reveals a consistent trend across all six shapes: the maximal von-Mises stress for shapes generated by our method is consistently lower than that of the shapes generated by the Diffusion-SDF method. This indicates that, under identical loading conditions, our method yields shapes with reduced occurrences of stress concentration at specific points, mitigating the risk of localized high-stress levels. Consequently, the shapes generated by our method demonstrate enhanced load-bearing capacity and are less susceptible to damage. These findings underscore the efficacy of our proposed approach.
}

\begin{figure}[t]
    \centering
    \includegraphics[width=0.8\linewidth]{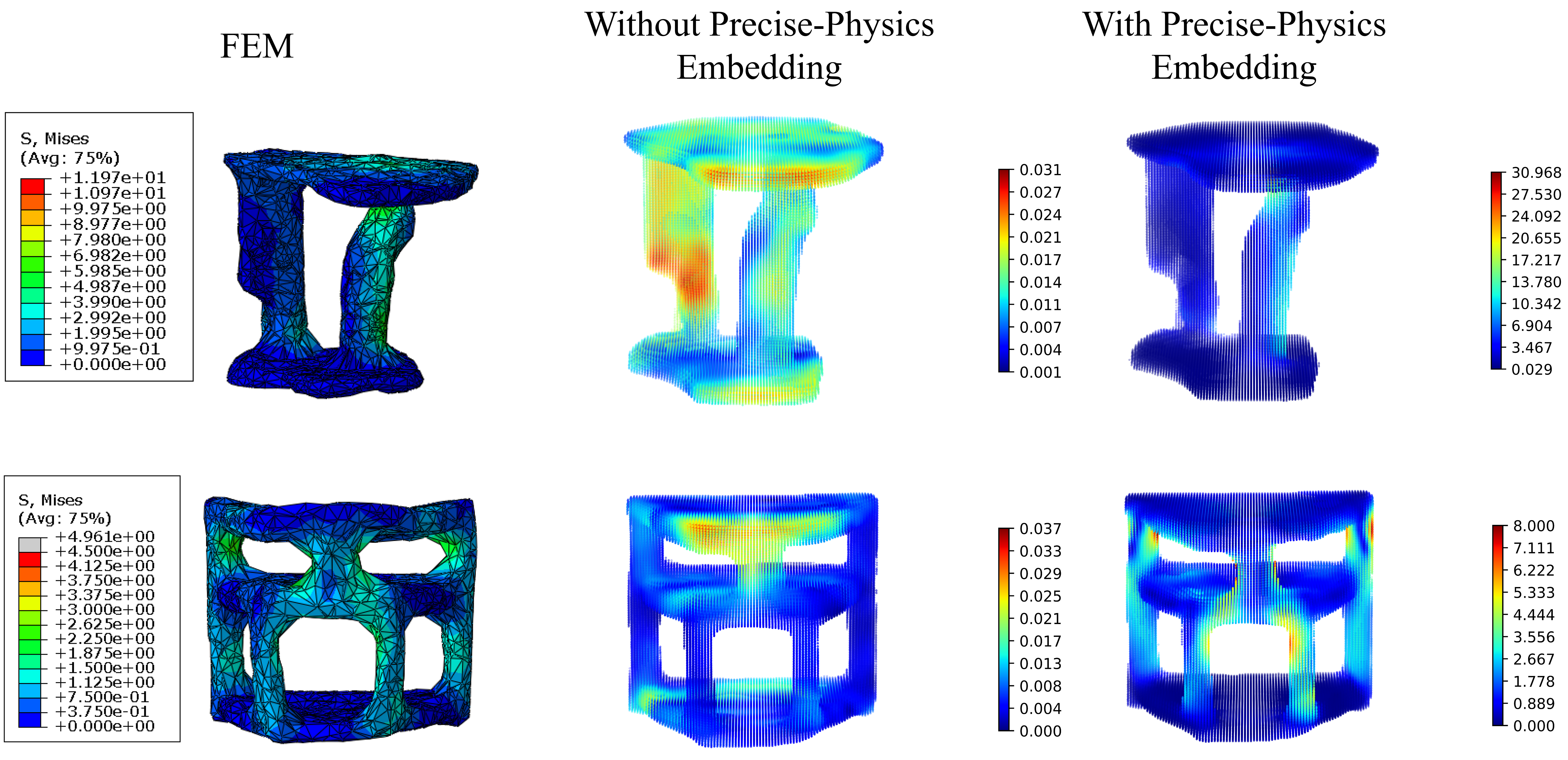}
    \caption{\textbf{The physics predictions for the tear drop table (upper) and the two-layered table (lower) provided by the FEM (left), the physical layer without precise-physics embedding, and the physical layer with precise-physics embedding.}}
    \label{fig:physics_results}
\end{figure}

{
Furthermore, in Fig.~\ref{fig:physics_results} we present the predicted von-Mises stress provided by our physics layer for tear drop table and two-layered table, alongside the corresponding results obtained from FEM simulations. A comparison between the two reveals that, although the predicted von-Mises stress values are not identical to those obtained from FEM, they still exhibit stress distributions consistent with physical laws. In contrast, the physics layer fails to provide accurate and reasonable physical predictions without employing precise physics embedding.
}

\subsection{Ablation Study}

\begin{figure}[t]
    \centering
    \includegraphics[width=0.95\linewidth]{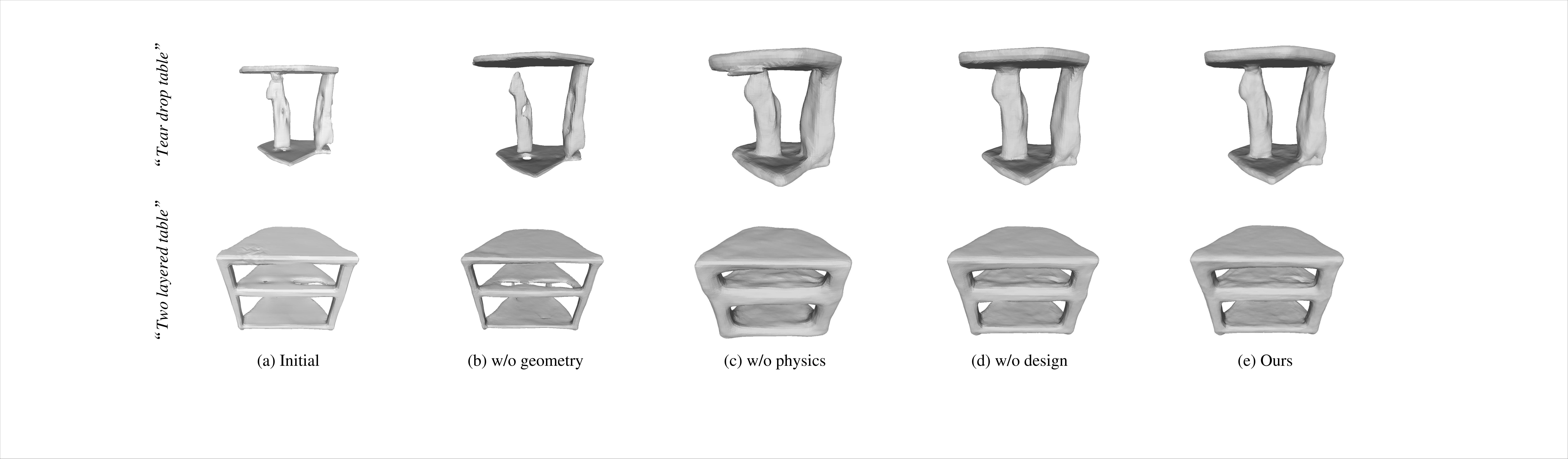}
    \caption{\textbf{Ablation study results.}}
    \label{fig:ablation}
\end{figure}

In this section, we conduct experiments on two 3D shapes to study the effect of different designs in our methods, including ``tear drop table'' and ``two layered table''. For comparison, we use the geometry generated from Diffusion-SDF as our initial geometry. \cref{fig:ablation} shows the ablation study results.

\noindent\textbf{Effect of geometry constraints.} After removing the geometry constraints, \cref{fig:ablation}(b) shows that the geometries become incomplete. Therefore, geometry constraints help the geometry optimization to keep shape information. 

\noindent\textbf{Effect of physics loss $\mathcal{L}_{pl}$.} By removing the physics loss, \cref{fig:ablation}(c) shows that the geometries contain some noise and are a bit fat. This will lead the force distribution on the geometries to be non-uniform. Since the physics loss can help the force distribution become more reasonable, it can help geometry become more smooth and practical.

\noindent\textbf{Effect of design loss $\mathcal{L}_{design}$.} By removing the design loss, \cref{fig:ablation}(d) shows that some regions of the geometries vary widely from the initial geometries, such as the left joint in the ``tear drop table''. This deviates from the visual preferences captured in the respective initial geometries.

\section{Conclusion}

In this paper, we have proposed a precise-physics-driven text-to-3D generation method to endow generated 3D shapes with precise physics perception capability. Our work reveals that the 3D shapes generated by existing text-to-3D generative methods do not conform to the laws of physics. Therefore, current generated 3D shapes are impractical in many real-world applications. To address this, we design a data-driven physics layer to learn precise physics information on demand when optimizing geometries. This guides our generated 3D shapes to satisfy both visual preferences and precise physics perception, opening up a new avenue for text-to-3D generation.

\section*{Acknowledgements}

The authors would like to thank Ge Jin, Ryan Lau, Jian Cheng Wong and Chin Chun Ooi for their help and discussions on this project.

%
%
\bibliographystyle{splncs04}
\bibliography{main}

\renewcommand\thesection{\Alph{section}}







\clearpage
\setcounter{section}{0}

\begin{center}
    \Large\textbf{Precise-Physics Driven Text-to-3D Generation \\
--Supplementary Material--}
\end{center}

In this supplementary document, we first provide the background knowledge of linear elastic solid mechanics in \cref{sec:equation_detail}. 
In \cref{sec:imp_detail}, we deep dive into the mechanism of Phy3DGen, revealing its effectiveness in the loss function. 
Then, we present additional experimental analysis in \cref{sec:exp_analysis}, including more ablation analysis and generalization of our method. 
Finally, we discuss our method's limitation and future work in \cref{sec:limitation_future}.

\section{Linear Elastic Solid Mechanics}\label{sec:equation_detail}
Though we introduced the governing functions of the linear elastic solid mechanics in Section 3.3, they may not be easy for the audience to understand without mechanical backgrounds. Henceforth, this section provides essential background knowledge of linear elastic solid mechanics. 

Based on the assumption of isotropic homogenous material, both the Cauchy stress tensor and the strain tensor are symmetric, \ie, $\epsilon_{12} = \epsilon_{21}$, $\epsilon_{13} = \epsilon_{31}$, $\epsilon_{23} = \epsilon_{32}$, $\sigma_{12} = \sigma_{21}$, $\sigma_{13} = \sigma_{31}$, and $\sigma_{23} = \sigma_{32}$. In this way, the governing equations can be expanded as follows, which are called the strain-displacement equations, the constitutive equations, and the equilibrium equations:
\begin{itemize}
    \item \textbf{Strain-displacement equations}: This set of equations establishes the relationship between the displacement field and the strain field, which are described as:

        \begin{equation}\label{pde:strain_displacement_x}
            \epsilon_{11}(\textbf{x}) = \frac{\partial u_1(\textbf{x})}{\partial x_1},
        \end{equation}
        
        \begin{equation}\label{pde:strain_displacement_y}
            \epsilon_{22}(\textbf{x}) = \frac{\partial u_2(\textbf{x})}{\partial x_2}, 
        \end{equation}

        \begin{equation}\label{pde:strain_displacement_z}
            \epsilon_{33}(\textbf{x}) = \frac{\partial u_3(\textbf{x})}{\partial x_3},
        \end{equation}

        \begin{equation}\label{pde:strain_displacement_xy}
            \epsilon_{12}(\textbf{x}) = \frac{\partial u_1(\textbf{x})}{\partial x_2} + \frac{\partial u_2(\textbf{x})}{\partial x_1},
        \end{equation}

        \begin{equation}\label{pde:strain_displacement_xz}
            \epsilon_{13}(\textbf{x}) = \frac{\partial u_1(\textbf{x})}{\partial x_3} + \frac{\partial u_3(\textbf{x})}{\partial x_1},
        \end{equation}

        \begin{equation}\label{pde:strain_displacement_yz}
            \epsilon_{23}(\textbf{x}) = \frac{\partial u_2(\textbf{x})}{\partial x_3} + \frac{\partial u_3(\textbf{x})}{\partial x_2},
        \end{equation}

    \item {\textbf{Constitutive equations}: This set of equations mainly establishes the relationship between the strain field and the stress field: 
        \begin{equation}\label{pde:stress_displacement_x}
            \sigma_{11}(\textbf{x}) = (\lambda + 2 \mu) \cdot \epsilon_{11}(\textbf{x}) +  \lambda \cdot \epsilon_{22}(\textbf{x}) + \lambda \cdot \epsilon_{33}(\textbf{x}),
        \end{equation}
        
        \begin{equation}\label{pde:stress_displacement_y}
            \sigma_{22}(\textbf{x}) = (\lambda + 2 \mu) \cdot \epsilon_{22}(\textbf{x}) +  \lambda \cdot \epsilon_{11}(\textbf{x}) + \lambda \cdot \epsilon_{33}(\textbf{x}),
        \end{equation}

        \begin{equation}\label{pde:stress_displacement_z}
            \sigma_{33}(\textbf{x}) = (\lambda + 2 \mu) \cdot \epsilon_{33}(\textbf{x}) +  \lambda \cdot \epsilon_{22}(\textbf{x}) + \lambda \cdot \epsilon_{33}(\textbf{x}),
        \end{equation}
        
        \begin{equation}\label{pde:stress_displacement_xy}
            \sigma_{12}(\textbf{x}) =  \mu \cdot \epsilon_{12}(\textbf{x}),
        \end{equation}

        \begin{equation}\label{pde:stress_displacement_xz}
            \sigma_{13}(\textbf{x}) =  \mu \cdot \epsilon_{13}(\textbf{x}),
        \end{equation}

        \begin{equation}\label{pde:stress_displacement_yz}
            \sigma_{23}(\textbf{x}) =  \mu \cdot \epsilon_{23}(\textbf{x}),
        \end{equation}
        where $\lambda = \frac{E \nu}{(1+\nu)(1-2\nu)}$ and $\mu = \frac{E}{2(1 + \nu)}$ are Lame parameters, and $E$ and $\nu$ are Young's module and Possion ratio, respectively.
    }
    \item \textbf{Equilibrium equations}: This set of equations aims to describe the equilibrium relationship of forces, \ie,
        \begin{equation}
            \frac{\partial \sigma_{11}(\textbf{x})}{\partial x_1} + \frac{\partial \sigma_{xy}(\textbf{x})}{\partial x_2} + \frac{\partial \sigma_{13}(\textbf{x})}{\partial x_3} + F_1(\textbf{x}) = 0,
        \end{equation}
        
        \begin{equation}
            \frac{\partial \sigma_{12}(\textbf{x})}{\partial x_1} + \frac{\partial \sigma_{22}(\textbf{x})}{\partial x_2} + \frac{\partial \sigma_{23}(\textbf{x})}{\partial x_3} + F_2(\textbf{x}) = 0,
        \end{equation}

        \begin{equation}
            \frac{\partial \sigma_{13}(\textbf{x})}{\partial x_1} + \frac{\partial \sigma_{23}(\textbf{x})}{\partial x_2} + \frac{\partial \sigma_{33}(\textbf{x})}{\partial x_3} + F_3(\textbf{x}) = 0,
        \end{equation}
        where $\textbf{F} = (F_1(\textbf{x}),F_2(\textbf{x}),F_3(\textbf{x}))$ is the already known force field, \eg, gravitational field. For simplicy, we assume $\textbf{F} = \textbf{0}$ here.
    \end{itemize}

    Moreover, the boundary conditions can be described as follows:
    \begin{itemize}
        \item \textbf{Boundary Conditions}: The first is the surface stress boundary condition. It describes that the force should be at equilibrium at the boundary $\Gamma_f$:
        \begin{equation}
            \sigma_{11}(\textbf{x}) \cdot n_1(\textbf{x}) + \sigma_{12}(\textbf{x}) \cdot n_2(\textbf{x}) + \sigma_{13}(\textbf{x}) \cdot n_3(\textbf{x}) = F_{n_1}(\textbf{x}),
        \end{equation}

        \begin{equation}
            \sigma_{12}(\textbf{x}) \cdot n_1(\textbf{x}) + \sigma_{22}(\textbf{x}) \cdot n_2(\textbf{x}) + \sigma_{23}(\textbf{x}) \cdot n_3(\textbf{x}) = F_{n_2}(\textbf{x}),
        \end{equation}

        \begin{equation}
            \sigma_{13}(\textbf{x}) \cdot n_1(\textbf{x}) + \sigma_{23}(\textbf{x}) \cdot n_2(\textbf{x}) + \sigma_{33}(\textbf{x}) \cdot n_3(\textbf{x}) = F_{n_3}(\textbf{x}),
        \end{equation}
        where $\textbf{n} = \left( n_1(\textbf{x}),n_2(\textbf{x}) n_3(\textbf{x}) \right)$ is the normal vector of a point $\textbf{x}$ on the boundary $\Gamma_t$, $\bar{\textbf{F}} = (F_{n_1}(\textbf{x}),F_{n_2}(\textbf{x}),F_{n_3}(\textbf{x}))$ is the force at the boundary point $\textbf{x} \in \Gamma_t$ (decomposed in the direction of the normal vector). The second one is called displacement boundary conditions, which means that the displacements of boundary points $\textbf{x} \in \Gamma_u$ are already known, \ie,
        \begin{equation}
            u_1(\textbf{x}) = \bar{u}_1(\textbf{x}),
        \end{equation}
   
        \begin{equation}
            u_2(\textbf{x}) = \bar{u}_2(\textbf{x}),
        \end{equation}

        \begin{equation}
            u_3(\textbf{x}) = \bar{u}_3(\textbf{x}),
        \end{equation}
        where $\bar{\textbf{u}} = (u_1(\textbf{x}),u_2(\textbf{x}),u_3(\textbf{x}))$ is the already known displacement.
\end{itemize}

\section{The Mechanism of Phy3DGen}\label{sec:imp_detail}
In this section, we delve deeper into the effectiveness of different loss components and provide further insights into our implementation details. In Section 3.3, we introduced three groups of loss components: physics loss components ($L_{pde}$, $L_{bc}$, and $L_{fem}$), geometry loss components ($L_{gc}$, $L_{vr}$, and $L_{eikonal}$), and the design loss component $L_{design}$, which utilizes physical information to guide geometry optimization. In order to improve the sturdiness of the geometry and maintain its shape, we proposed the following optimization strategy. Initially, our implementation does not incorporate the design loss component. Instead, we rely on geometry loss components to facilitate the optimization, leading to an increase in the geometry volume. As the geometry reaches the target volume $M_{v}$, we limit the growth only to the high-stress regions based on the guidance of physics information. This is accomplished through a combination of $L_{design}$ and $L_{vr}$. Thus, in our implementation, we use the following simplified version:
\begin{equation}\label{combine}
    L_{combine} = \delta \left( \max \{ \frac{\sum_{\textbf{x} \in \mathcal{D}_{\Omega^{+}}} \hat{\rho} - M_{v}}{|\mathcal{D}_{\Omega^{+}}|}, 0 \} \right) \cdot \left( \frac{\sum_{\textbf{x} \in \mathcal{D}_{\Omega^{+}}}\{(\hat{\sigma}_{vm,max} - \hat{\sigma}_{vm}) \hat{\rho} \}}{|\mathcal{D}_{\Omega^{+}}|} \right),
\end{equation}
where $\delta(\cdot)$ represents the Dirichlet function, and $\hat{\sigma}_{vm,max} = \max_{\textbf{x} \in \Omega^{+}} \{ \hat{\sigma}_{vm} \}$ is detached during the training process. This combination serves a specific purpose: if the geometry surpasses the target volume, the design loss aids in limiting the growth or even removing the regions characterized by low von-Mises stress. Such regions are typically non-load-bearing and contribute minimally to the structure's functionality. This evolutionary approach offers several advantages. Firstly, it eliminates geometry defects inherited during the 3D model generation. Secondly, it fosters the development of a more sturdy geometry by aligning with principles of physics, thus enhancing the overall realism of the design.

\section{Experimental Analysis}\label{sec:exp_analysis}

In this section, we conducted additional ablation studies to demonstrate the effectiveness of our method. In addition, we show the generalization of our method on another 3D diffusion model, Shap-E~\cite{jun2023shap}.

\subsection{Additional Ablation Study}

\begin{figure}
    \centering
    \includegraphics[width=\linewidth]{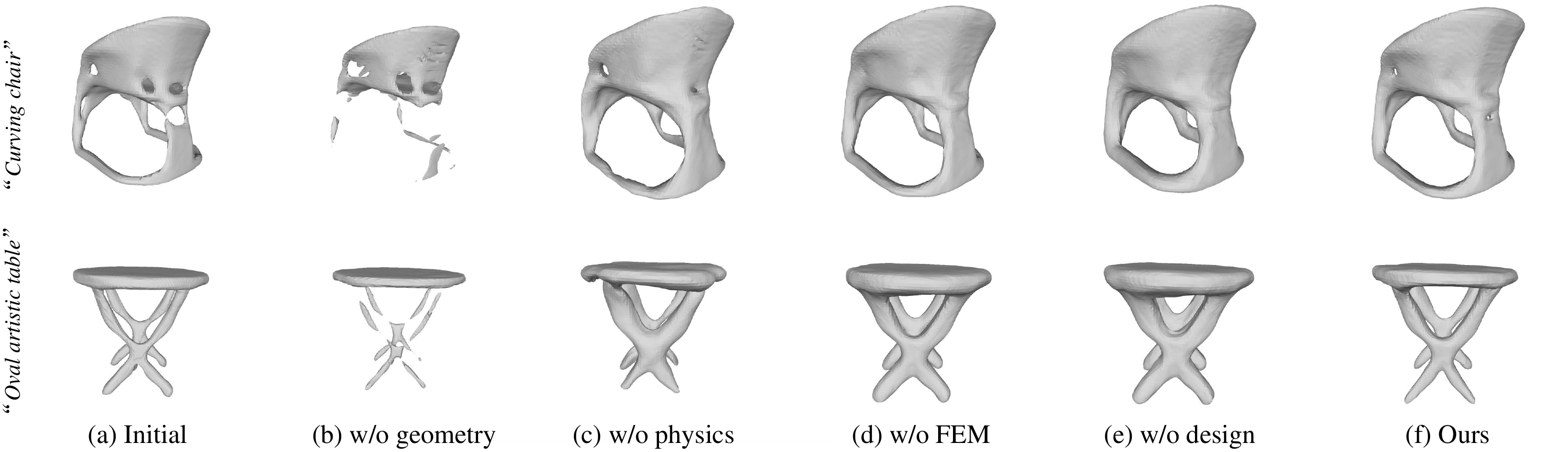}
    \caption{Additional ablation study results demonstrated that our proposed method eliminates the geometry defects introduced from Diffusion-SDF generation, strengthening the weak joints to make the optimized geometry more realistic and sturdy. The combined loss components work in harmony to ensure the shape integrity and sturdiness of the geometry. Hence, removing any lost component will result in degrading 3D shapes.}
    \label{fig:supp_ablation}
\end{figure}

Here, we conducted additional experiments on the ``curving chair'' and ``oval artistic table'' to study the effect of different designs on our proposed method. The results are reported in \cref{fig:supp_ablation}. We observe that the initial geometries generated by Diffusion-SDF~\cite{li2023diffusion} contained holes or weak joints, as shown in \cref{fig:supp_ablation}(a). In contrast, by incorporating physics into our method, our generated shapes can significantly alleviate these defects. As for our different designs, without geometry constraints, the geometries cannot maintain the shape information. Without physics loss, the geometries become rough and uneven. Without FEM embedding, some details cannot be preserved. Without design loss, the geometries become fatter. These comparison experiments effectively demonstrate that our different designs facilitate our method to generate high-quality 3D models. 

\subsection{Discussion on Generalizability}

\begin{figure}
    \centering
    \includegraphics[width=\linewidth]{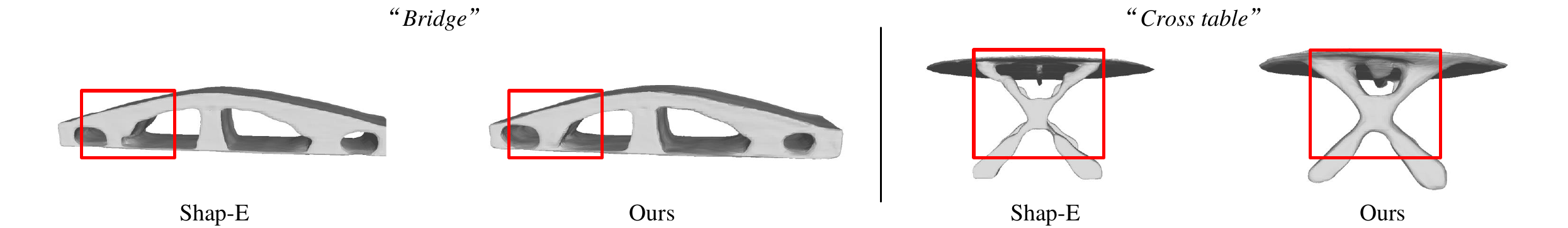}
    \caption{Our proposed method is not limited to only 3D shapes generated from Diffusion-SDF model, but also from the Shap-E model presented here.}
    \label{fig:shap-e}
\end{figure}

To investigate the generalization of our method, we employ Shap-E as our 3D generative model to synthesize initial geometries. As presented in the results in \cref{fig:shap-e}, when the generated 3D shapes have inherent defects, such as uneven surfaces or weak joints, our method can effectively eliminate these defects, resulting in more realistic 3D shapes. 

\section{Limitation and Future Work}\label{sec:limitation_future}

While our method can generate high-quality 3D shapes with the assistance of precise physics, there are still some limitations: 1) Efficiency. We use vanilla MLPs to parameterize our geometry and physics representations. This compromises the efficiency of our approach. Inspired by Instant-NGP~\cite{muller2022instant}, we can leverage multi-resolution hash encoding to parameterize these two representations to improve our efficiency. 2) Precision of Physics. The precision of our physics information depends in part on the number of sampling points, following the similar reason explained in \cite{chiu2022can}. In the future, we may increase the number of sampling points to boost the precision of physics information using more efficient geometry and physics representations.

\end{document}